\documentclass{article}

\usepackage[main, final]{neurips_2026}
\usepackage{listings}

\usepackage[utf8]{inputenc} 
\usepackage[T1]{fontenc}    
\usepackage{hyperref}       
\usepackage{url}            
\usepackage{booktabs}       
\usepackage{amsfonts}       
\usepackage{nicefrac}       
\usepackage{microtype}      
\usepackage{xcolor}  

\usepackage{natbib} 
\usepackage{amsmath}
\usepackage{algorithm}
\usepackage{algorithmic}

\usepackage[capitalize,noabbrev]{cleveref}

\usepackage[textsize=tiny]{todonotes}

\usepackage[T1]{fontenc}

\usepackage[utf8]{inputenc}

\usepackage{microtype}

\usepackage{inconsolata}

\usepackage{graphicx}

\def\BibTeX{{\rm B\kern-.05em{\sc i\kern-.025em b}\kern-.08em
    T\kern-.1667em\lower.7ex\hbox{E}\kern-.125emX}}

\title{Enhancing Small Language Models for Power Outage Report Generation via Minimum Risk Training}

\author{
  Hung Phan \\
  Iowa State University \\
  \texttt{hungphd@iastate.edu}
  \And
  Waqwoya Abebe \\
  Oak Ridge National Laboratory \\
  \texttt{abebewm@ornl.gov}
  \And
  Youssef Hussein \\
  University of Minnesota \\
  \texttt{husse408@umn.edu}
  \AND
  Supriya Chinthavali \\
  Oak Ridge National Laboratory \\
  \texttt{chinthavalis@ornl.gov}
  \And
  Dalton Lunga \\
  Oak Ridge National Laboratory \\
  \texttt{lungadd@ornl.gov}
  \And
  Ali Jannesari \\
  Iowa State University \\
  \texttt{jannesar@iastate.edu}
}

\begin{document}

\maketitle

\begin{abstract}
Minimum Risk Training (MRT) enables neural machine translation models to directly optimize sequence-level evaluation metrics instead of relying only on token-level maximum-likelihood objectives~\cite{001_shen2016minimumrisktrainingneural}. Although introduced a decade ago, recent work shows renewed potential for risk-based optimization in modern language models~\cite{014_yang-etal-2024-direct,016_jinnai-etal-2025-regularized}. We apply MRT to power outage report generation for the Outage Data Initiative Nationwide (ODIN), transforming heterogeneous reports into standardized XML compliant with CIM IEC 61968-3. Our MRT approach improves Qwen2.5-7B-Instruct overall accuracy from 16.20\% to 68.95\%, demonstrating the effectiveness of sequence-level optimization for domain-specific structured generation.
\end{abstract}

\section{Introduction}

Small Language Models (SLMs) have recently emerged as efficient alternatives to large language models (LLMs) for text and code generation, particularly when models can be specialized for a target task. Their smaller model sizes enable lower memory consumption, faster inference, and more practical deployment on limited computational resources. Recent studies show that carefully optimized SLMs can achieve competitive performance on both natural-language and code-generation tasks~\cite{007_hsieh-etal-2023-distilling,006_hasan2026assessingsmalllanguagemodels}. However, this efficiency introduces an important trade-off: smaller models generally have less model capacity and may underperform larger models when solving complex or general-purpose generation tasks. This motivates training strategies that improve the task-specific generation quality of SLMs without sacrificing their computational advantages.

One such application is power outage report generation. The Outage Data Initiative Nationwide (ODIN)\footnote{\url{https://odin.ornl.gov/}} standardizes the exchange of power outage information to support response, restoration, and interoperability among utilities and other stakeholders. In this work, we consider the transformation of heterogeneous, non-standard outage reports into standardized XML reports following the Common Information Model (CIM) IEC 61968-3 specification. Unlike open-ended text generation, this task requires a model to preserve the outage information while simultaneously producing a strictly structured output with the appropriate XML elements and organization. Such domain-specific requirements can be challenging for general-purpose language models and particularly for computationally efficient SLMs with limited model capacity.

Recent research suggests that classical machine-learning techniques can still complement modern pretrained models when they provide task-specific information that the newer model does not capture effectively. Phan and Jannesari revisit Statistical Machine Translation (SMT), a classical machine-translation technique, to improve modern BERT-based code-search models~\cite{002_phansmt}. Their Oracle4CS approach uses SMT to translate natural-language queries into a compact Abstract Syntax Tree representation, called ASTSum, and incorporates the predicted representation into modern domain specific code-search pipelines. 
This work's result demonstrates that revisiting classical optimization and learning techniques can provide complementary signals to modern neural models, motivating us to investigate a similar direction for improving domain-specific SLM generation.

Following this motivation, we leverage Minimum Risk Training (MRT), a sequence-level optimization method originally developed for machine translation~\cite{001_shen2016minimumrisktrainingneural} for power outage (PO) report generation. Unlike conventional maximum-likelihood training, MRT directly minimizes the expected task-specific risk over multiple candidate outputs, allowing generation models to optimize sequence-level evaluation objectives. Its effectiveness has subsequently been demonstrated beyond conventional translation, including improving the naturalness and diversity of referring-expression generation~\cite{009_panagiaris-etal-2020-improving} and mitigating exposure bias in domain-specific biomedical translation~\cite{010_saunders-byrne-2020-addressing}; more recent work further demonstrates that MRT can be used to investigate and optimize neural sequence-level evaluation metrics~\cite{011_yan2023bleurtuniversaltranslationsanalysis}. Motivated by these results, we propose \textbf{PO-MRT}, an MRT-based fine-tuning pipeline for power outage report generation. PO-MRT samples multiple candidate reports from Qwen2.5-7B-Instruct, evaluates their sequence-level risk using sentence-level BLEU, and directly optimizes the expected risk through parameter-efficient fine-tuning, enabling the SLM to learn from complete generated reports rather than relying solely on token-level likelihood.

\section{Power Outage Report Generation}
\label{sec:background}

The Power Outage Tracker considered in this work is part of the Outage Data Initiative Nationwide (ODIN) system developed at Oak Ridge National Laboratory (ORNL).\footnote{\url{https://odin.ornl.gov/outagemap/index.html}} Its objective is to convert heterogeneous power outage information into a unified representation that follows the Common Information Model (CIM) IEC 61968-3 specification~\cite{IEC61968-3:2021}. In practice, outage information can be reported in substantially different forms, including incomplete or non-standard XML and JSON records as well as natural-language descriptions. Consequently, report generation requires more than producing fluent text: the generated output must preserve the information contained in the source report while satisfying the structural requirements of the target XML standard. This combination of semantic preservation and strict structural compliance makes power outage report generation a domain-specific structured generation problem.



\section{Motivation Example}
\label{sec:motivation}

Power outage report generation differs from conventional text generation because
the model must simultaneously preserve information from a heterogeneous input
and satisfy a strict XML schema. Consider the simplified example below. A utility
initially reports an outage using a non-standard JSON representation:

\begin{lstlisting}[
    basicstyle=\ttfamily\scriptsize,
    frame=single,
    caption={Example of a non-standard power outage report.},
    label={lst:motivation-input}
]
{
  "outage_id": "A10482",
  "customers_affected": 742,
  "latitude": "35.9606",
  "longitude": "-83.9207",
  "reported_time": "2026-07-18T14:20:00Z",
  "restore_by": "2026-07-18T17:30:00Z",
  "message": "Crews are responding to an outage
              affecting 742 customers."
}
\end{lstlisting}

The same information must be transformed into a standardized XML report whose
content and hierarchy conform to the ODIN representation of CIM IEC 61968-3:

\begin{lstlisting}[
    basicstyle=\ttfamily\scriptsize,
    frame=single,
    caption={Corresponding standardized power outage report.},
    label={lst:motivation-output}
]
<?xml version="1.0" encoding="UTF-8"?>
<po:PubOutages
  xmlns:po="http://iec.ch/TC57/2014/PubOutages#">
  <po:Outage>
    <po:mRID>A10482</po:mRID>
    <po:customerCount>742</po:customerCount>
    <po:outageKind>outageReported</po:outageKind>
    <po:reportedStartTime>
      2026-07-18T14:20:00Z
    </po:reportedStartTime>
    <po:EstimatedRestorationTime>
      <po:ert>2026-07-18T17:30:00Z</po:ert>
    </po:EstimatedRestorationTime>
    <po:Incident>
      <po:Location>
        <po:latitude>35.9606</po:latitude>
        <po:longitude>-83.9207</po:longitude>
      </po:Location>
    </po:Incident>
  </po:Outage>
</po:PubOutages>
\end{lstlisting}

This transformation exposes two complementary requirements.
First, the generated report must achieve \emph{textual similarity}: outage-specific
information such as the identifier, number of affected customers, timestamps,
location, and other semantic content must remain consistent with the source
report. Second, it must achieve \emph{tag similarity}: the generated XML must use
the appropriate CIM tags, nesting relationships, namespaces, and opening and
closing elements. Satisfying only one requirement is insufficient. For example,
a model may correctly generate the value \texttt{742} but place it under an
incorrect XML element, while another output may reproduce the correct XML
structure but hallucinate an incorrect customer count or restoration time.

Consequently, two generated reports with similar token-level likelihood can
have substantially different usefulness. A report with high textual similarity
may still violate the required XML organization, whereas a structurally valid
report may contain incorrect outage information. This mismatch is particularly
challenging for a small language model, since conventional token-level
fine-tuning does not explicitly distinguish between complete candidate reports
according to their report-level quality. This example motivates the use of
Minimum Risk Training, where multiple complete candidate reports can be
evaluated according to task-specific generation quality and the model can be
optimized toward candidates with lower sequence-level risk.

\begin{algorithm}
\caption{QLoRA Fine-tuning with Minimum Risk Training (MRT) using Sentence BLEU}
\label{alg:mrt-bleu}
\begin{algorithmic}[1]
\REQUIRE Training set $\mathcal{D}=\{(p^{(n)}, y^{(n)})\}_{n=1}^{N}$ with prompt $p$ and reference response $y$; base causal LM $f_\theta$; tokenizer $\mathcal{T}$; max length $L_{\max}$.
\REQUIRE Generation hyperparameters: number of candidates $K$, temperature $T$, nucleus $p$, max new tokens $L_{\text{gen}}$.
\REQUIRE Risk temperature $\tau > 0$; risk $\Delta(\hat{y}, y) = 1-\mathrm{BLEU}(\hat{y}, y)$.
\ENSURE LoRA adapter parameters (and optionally other trainable parameters) updated to minimize expected risk.

\vspace{0.3em}
\STATE \textbf{(Data preprocessing)} FOR each example $(p,y)\in \mathcal{D}$:
\STATE \hspace{1em} Format prompt: $\tilde{p}\leftarrow \texttt{" Instruction:\textbackslash n"} \,\Vert\, p \,\Vert\, \texttt{"\textbackslash n\textbackslash n Response:\textbackslash n"}$
\STATE \hspace{1em} Tokenize: $x_p\leftarrow \mathcal{T}(\tilde{p}),\;\; x_y\leftarrow \mathcal{T}(y \Vert \texttt{EOS})$
\STATE \hspace{1em} Truncate to $L_{\max}$ prioritizing keeping $x_y$ (drop prefix of $x_p$ if needed)
\STATE \hspace{1em} Build input: $x \leftarrow x_p \Vert x_y$
\STATE \hspace{1em} Build labels (response-only supervision mask): $\ell \leftarrow \underbrace{(-100,\dots,-100)}_{|x_p|}\Vert x_y$

\vspace{0.3em}
\STATE \textbf{(Model setup)} Load $f_\theta$ in 4-bit (NF4) and attach LoRA adapters to projection modules; enable gradient checkpointing; disable cache during training.

\vspace{0.3em}
\FOR{each training step with mini-batch $\{(x^{(b)},\ell^{(b)})\}_{b=1}^{B}$}
    \STATE $L \leftarrow 0$
    \FOR{$b=1$ to $B$}
        \STATE Identify response start: $m \leftarrow \min\{t:\ell^{(b)}_t \neq -100\}$ \COMMENT{prompt length}
        \STATE Prompt tokens: $x^{(b)}_{1:m}$; reference response tokens: $x^{(b)}_{\text{ref}}\leftarrow \{ \ell^{(b)}_t : \ell^{(b)}_t \neq -100\}$

        \vspace{0.2em}
        \STATE \textbf{(Candidate generation; no grad)} Sample $K$ candidates:
        \STATE \hspace{1em} $\{\hat{x}^{(b)}_k\}_{k=1}^{K} \leftarrow \mathrm{Generate}\!\left(f_\theta, x^{(b)}_{1:m}; K, T, p, L_{\text{gen}}\right)$
        \STATE \hspace{1em} Each $\hat{x}^{(b)}_k$ includes the prompt prefix; extract candidate response $\hat{x}^{(b)}_{k,\text{resp}} \leftarrow \hat{x}^{(b)}_k[m{+}1:\,]$

        \vspace{0.2em}
        \STATE \textbf{(Risk computation; no grad)} FOR each $k$:
        \STATE \hspace{1em} $\mathrm{risk}^{(b)}_k \leftarrow 1 - \mathrm{BLEU}\!\left(\mathcal{T}^{-1}(x^{(b)}_{\text{ref}}), \mathcal{T}^{-1}(\hat{x}^{(b)}_{k,\text{resp}})\right)$

        \vspace{0.2em}
        \STATE \textbf{(Candidate scoring; with grad)} FOR each $k$, compute log-prob of the \emph{response} under $f_\theta$:
        \STATE \hspace{1em} $s^{(b)}_k \leftarrow \log p_\theta(\hat{x}^{(b)}_{k,\text{resp}} \mid x^{(b)}_{1:m})
        = \sum_{t=m}^{|\hat{x}^{(b)}_k|-1} \log p_\theta\!\left(\hat{x}^{(b)}_{k,t+1}\mid \hat{x}^{(b)}_{k,1:t}\right)$

        \vspace{0.2em}
        \STATE \textbf{(Soft candidate distribution)} $q^{(b)} \leftarrow \mathrm{softmax}\!\left(\frac{1}{\tau}[s^{(b)}_1,\dots,s^{(b)}_K]\right)$
        \STATE \textbf{(Expected risk)} $L_b \leftarrow \sum_{k=1}^{K} q^{(b)}_k \cdot \mathrm{risk}^{(b)}_k$
        \STATE $L \leftarrow L + L_b$
    \ENDFOR
    \STATE $L \leftarrow \frac{1}{B}L$
    \STATE Update $\theta$ (LoRA parameters) by backpropagating $\nabla_\theta L$
\ENDFOR
\end{algorithmic}
\end{algorithm}
\section{Minimum Risk Training for Power Report Generation}
\label{sec:approach}

\subsection{Why Minimum Risk Training}
\label{sec:why-mrt}

Minimum Risk Training (MRT) was originally proposed for neural machine translation to address a limitation of conventional maximum-likelihood training: the model is optimized at the token level, while translation quality is typically evaluated at the sequence level~\cite{001_shen2016minimumrisktrainingneural}. Instead of maximizing only the likelihood of the reference output, MRT generates a set of candidate sequences, evaluates each candidate using a task-specific loss, and minimizes the expected risk over these candidates.

A key advantage of MRT is that the evaluation function does not need to be differentiable, allowing sequence-level metrics such as BLEU to be incorporated directly into training. Because enumerating all possible output sequences is infeasible, the original MRT algorithm approximates the output space through sampling and assigns higher training preference to candidates with lower sequence-level risk. This makes MRT particularly suitable for structured generation tasks, where the quality of the complete output is more important than the correctness of individual tokens in isolation.

\subsection{MRT Fine-tuning for XML Report Generation}
\label{sec:mrt-xml}

We adapt MRT to fine-tune Qwen2.5-7B-Instruct for standardized power outage XML generation. Each training example contains a non-standard outage report as the prompt and a corresponding standardized XML report as the reference. The model is loaded with 4-bit quantization and optimized using LoRA adapters, allowing sequence-level MRT to be incorporated into parameter-efficient fine-tuning.

For each prompt, the current model samples $K$ candidate XML reports. Each candidate is compared with the reference XML using sentence-level BLEU, and its risk is defined as one minus the BLEU score. Candidates with higher BLEU therefore receive lower risk. The model then computes the likelihood of each candidate and forms a normalized distribution over the sampled reports. The training objective minimizes the expected risk across all candidates, encouraging the model to assign higher probability to XML outputs that more closely match the standardized reference.

Different from the original MRT work, which applies minimum-risk optimization to neural machine translation~\cite{001_shen2016minimumrisktrainingneural}, our approach integrates MRT into the fine-tuning process of a pretrained causal language model. This allows the model to learn from the quality of complete generated XML reports rather than relying only on token-level supervision.
\section{Experiment}
\label{sec:experiment}

\begin{table}[t]
\vspace{-0.05in}
\caption{Performance comparison across training objectives and baselines.}
\label{tab:config_results}
\centering
\small
\begin{tabular}{lccc}
\toprule
\textbf{Configuration} & \textbf{Overall} & \textbf{Text Acc/} & \textbf{Tag Acc.} \\
\midrule
Qwen2.5-7B-Instruct      & 16.20\% & 28.85\% & 3.56\% \\
MRT over Qwen2.5-7B-Instruct (Ours)       & 68.95\% & 50.73\% & 87.17\% \\
\bottomrule
\end{tabular}
\vspace{-0.10in}
\end{table}

\subsection{Experimental Setup}

\noindent\textbf{Dataset.}
We select 1,000 power outage report entities for our experiments. Each data point contains a non-standard outage report and its corresponding standardized XML report following the CIM IEC 61968-3 specification. These paired reports are used to fine-tune and evaluate Qwen2.5-7B-Instruct with and without Minimum Risk Training.

\noindent\textbf{Evaluation Metrics.}
We evaluate generated reports from two complementary perspectives: textual similarity and XML tag similarity. \emph{Text Accuracy} measures the similarity between the textual content of the generated report and the reference report. \emph{Tag Accuracy} measures structural similarity by extracting the XML tag sequences from both reports and calculating the common sequence of tags between them. The \emph{Overall Accuracy} is computed as the average of Text Accuracy and Tag Accuracy, allowing the evaluation to consider both semantic content and XML structural correctness.

\noindent\textbf{Hardware.}
All experiments are conducted on a server equipped with an NVIDIA RTX 3090 Ti GPU and an Intel Core i9 processor. The same hardware environment is used for both the baseline and MRT fine-tuning experiments.

\subsection{Results}

Table~\ref{tab:config_results} compares the original Qwen2.5-7B-Instruct model with our MRT-based fine-tuning approach. The baseline achieves only 16.20\% overall accuracy, including 28.85\% Text Accuracy and 3.56\% Tag Accuracy. Applying MRT substantially improves the overall accuracy to 68.95\%, representing a gain of 52.75 percentage points.

The largest improvement is observed in XML structural correctness, where Tag Accuracy increases from 3.56\% to 87.17\%. Text Accuracy also improves from 28.85\% to 50.73\%. These results indicate that sequence-level MRT is particularly effective for learning the structured XML representation required by power outage reports while simultaneously improving preservation of the report's textual content.
\section{Related Work}
\label{sec:related}

\noindent\textbf{Applications of MRT in Machine Learning Optimization.}
Beyond its original application to neural machine translation, Minimum Risk Training has been explored for several sequence-generation problems. Panagiaris et al.~\cite{009_panagiaris-etal-2020-improving} combine MRT with maximum-likelihood training for referring expression generation and show that sequence-level optimization can improve the naturalness and diversity of generated descriptions compared with reinforcement-learning-based alternatives. Saunders and Byrne~\cite{010_saunders-byrne-2020-addressing} extend MRT to document-level optimization for biomedical machine translation, using document-level evaluation metrics to reduce exposure-bias effects during domain-specific fine-tuning. More recently, Yan et al.~\cite{011_yan2023bleurtuniversaltranslationsanalysis} use MRT as a mechanism for analyzing neural evaluation metrics such as BLEURT and BARTScore, revealing robustness issues in these metrics and showing that additional token-level constraints can improve their behavior. Together, these works demonstrate that MRT provides a flexible framework for incorporating task-level quality measures directly into model optimization.

\noindent\textbf{Risk-Based Optimization for Large Language Models.}
Although direct applications of MRT to modern LLM fine-tuning remain relatively limited, recent work has increasingly explored closely related sequence-level and risk-based optimization strategies. Wang et al.~\cite{012_wang2023esrlefficientsamplingbasedreinforcement} study efficient sampling-based reinforcement learning for sequence generation and include MRT as an important sequence-level optimization baseline, while also extending their evaluation to RLHF for large language models. Minimum Bayes Risk (MBR) decoding provides another closely related direction: instead of updating model parameters, MBR generates multiple candidates and selects the output with the lowest expected risk. Suzgun et al.~\cite{013_suzgun2022followwisdomcrowdeffective} apply MBR decoding to open-ended generation tasks including summarization, data-to-text generation, translation, and style transfer. Yang et al.~\cite{014_yang-etal-2024-direct} further use MBR-generated preferences with Direct Preference Optimization to transfer MBR improvements into multilingual language models, reducing the need for expensive MBR decoding during inference. Multi-prompt MBR has also been proposed to improve candidate diversity for instruction-tuned LLMs~\cite{015_heineman2024improvingminimumbayesrisk}, while MBR-BoN incorporates an MBR objective into Best-of-$N$ sampling to reduce reward hacking in LLM alignment and to construct improved preference data for DPO~\cite{016_jinnai-etal-2025-regularized}. These studies suggest that expected-risk principles remain relevant to modern LLMs and motivate our investigation of MRT directly within the fine-tuning process.

\noindent\textbf{Potential Applications to Data Science.}
Risk-based optimization may also be useful for domain-specific data-science applications in which generated outputs must satisfy complex spatial, temporal, or structural constraints. Recent energy research applies federated learning and privacy-preserving techniques to short-term electricity load forecasting, highlighting the need for specialized learning approaches in data-sensitive power-system applications~\cite{017_Waqwoya}. In geospatial data science, Dias et al.~\cite{018_dias2026emergingflexibledesignsgeospatial} investigate flexible multimodal foundation-model architectures for Earth observation, emphasizing trade-offs among modality alignment, architectural flexibility, and downstream performance. Tran et al.\ propose GeoGNN, a two-tower graph neural network that combines temporal representations with geographic adjacency information for geolocating electricity-consumption time series, demonstrating the value of explicitly incorporating structural relationships into learning systems~\cite{019_tran2026geognn}. Large multimodal resources such as MOSAIC-CONUS further provide paired multi-temporal Earth-science data for developing domain-specific learning systems~\cite{020_osti_3005125}. These directions suggest that future domain-specific LLM optimization may benefit not only from sequence-level objectives such as MRT, but also from complementary techniques such as graph neural networks that explicitly encode relationships and structures that are difficult to capture through text generation alone.
\section{Conclusion}
\label{sec:conclusion}

This work demonstrates that Minimum Risk Training can effectively improve a small language model for structured power outage report generation while requiring only modest hardware through parameter-efficient fine-tuning. The strong improvement in XML generation accuracy suggests that MRT is a practical approach for domain-specific SLM optimization and may also benefit other data-science applications with structured or specialized outputs. Future work will investigate how different definitions of sequence-level \emph{risk} should be designed for different domains, including risks based on structure, semantics, spatial relationships, or other task-specific constraints. Replication package is available at here\footnote{https://tinyurl.com/583bpzry}.
\newpage
\bibliographystyle{plainnat}
\bibliography{refs}







\end{document}